\documentclass[dvipsnames,format=sigconf]{acmart}

\usepackage{subfig}
\usepackage[noend]{algpseudocode}
\usepackage{algorithm}
\usepackage{fancyvrb}
\usepackage{xfrac}
\usepackage{listings}
\usepackage{tikz}
\usetikzlibrary{positioning}
\usepackage[normalem]{ulem} % https://tex.stackexchange.com/questions/23711/strikethrough-text
\usepackage{enumitem}

\makeatletter
\newcommand\anonacm[1]{%
\if@ACM@anonymous
    \textcolor{red}{ANONYMIZED}%
  \else
    #1%
  \fi
}
\makeatother

\AtBeginDocument{%
  \providecommand\BibTeX{{%
    \normalfont B\kern-0.5em{\scshape i\kern-0.25em b}\kern-0.8em\TeX}}}

\copyrightyear{2025}
\acmYear{2025}
\setcopyright{acmlicensed}\acmConference[GECCO '25]{Genetic and Evolutionary Computation Conference}{July 14--18, 2025}{Malaga, Spain}
\acmBooktitle{Genetic and Evolutionary Computation Conference (GECCO '25), July 14--18, 2025, Malaga, Spain}
\acmDOI{10.1145/3712256.3726385}
\acmISBN{979-8-4007-1465-8/2025/07}
\begin{document}

\title{rEGGression: an Interactive and Agnostic Tool for the Exploration of Symbolic Regression Models}

\author{Fabrício Olivetti de França}
\orcid{0000-0002-2741-8736}
\affiliation{%
  \institution{Federal University of ABC}
  \city{Santo Andre}
  \state{São Paulo}
  \country{Brazil}
}
\email{folivetti@ufabc.edu.br}

\author{Gabriel Kronberger}
\orcid{0000-0002-3012-3189}
\affiliation{%
  \institution{University of Applied Sciences Upper Austria}  
  \city{Hagenberg}
  \state{Upper Austria}
  \country{Austria}
}
\email{gabriel.kronberger@fh-hagenberg.at}

%% \renewcommand{\shortauthors}{de Franca et al.}

\begin{abstract} % 200 words or less
Regression analysis is used for prediction and to understand the effect of independent variables on dependent variables.
Symbolic regression (SR) automates the search for non-linear regression models, delivering a set of hypotheses that balances accuracy with the possibility to understand the phenomena. Many SR implementations return a Pareto front allowing the choice of the best trade-off. However, this hides alternatives that are close to non-domination, limiting these choices. Equality graphs (e-graphs) allow to represent large sets of expressions compactly by  efficiently handling duplicated parts occurring in multiple expressions. The e-graphs allow to efficiently store and query all solution candidates visited in one or multiple runs of different algorithms and open the possibility to analyze much larger sets of SR solution candidates.
We introduce rEGGression, a tool using e-graphs to enable the exploration of a large set of symbolic expressions which provides querying, filtering, and pattern matching features creating an interactive experience to gain insights about SR models. The main highlight is its focus in the exploration of the building blocks found during the search that can help the experts to find insights about the studied phenomena. This is possible by exploiting the pattern matching capability of the e-graph data structure.
\end{abstract}




\maketitle

\section{Introduction}
% blah blah blah of what is symbolic regression
Regression analysis~\cite{gelman2007data} is an essential technique of machine learning and statistics with the objective of predicting unobserved data or, more importantly, gaining a better understanding of a given phenomena.
In many situations, a linear relationship is assumed to give a rough estimation of the effects of the independent variables on the dependent variable. A linear model is easy to understand, as the effect is constant, but it may fail to capture the non-linear relationship of the variables.
Using a non-linear model often requires expert knowledge about the phenomena and a manual and tedious process of exploring different recurrent patterns such as polynomials, exponential growth, among others. Another possibility is to use a \emph{generic} model such as generalized additive models~\cite{hastie2017generalized}, which still can be considered interpretable or using opaque models such as neural networks~\cite{bishop1994neural} and gradient boosting~\cite{chen2016xgboost}, which limits the analysis of the relationship of the variables.

Symbolic regression~\cite{Koza1992,kronberger2024} (SR) automates this process by searching for a mathematical model of a system that approximates a collection of data points of interest. This technique can provide a valuable help for the task of scientific discovery~\cite{kronberger2024,cranmerpysr, udrescu2020ai}. %,russeil2024multiview}.
% Use for inference
The main highlight of symbolic regression is the possibility of finding a non-linear model accurately describing the data that still can be used for inference, data exploration, and interpretability.%, unlike opaque models.

Although the main objective, when searching for a model, is to maximize accuracy, this comes with the cost of getting a complex model as a result. For this reason, many algorithms implement complexity control mechanisms such as penalization, model selection, maximum complexity constraint, and multi-objective optimization.

Multi-objective optimization~\cite{deb2002fast,smits2005pareto,kommenda2015complexity} transfers the decision on the best trade-off between accuracy and complexity to the user by presenting the whole Pareto front. Another alternative is to return multiple solutions that are semantically similar int the training set, but different outside that boundaries~\cite{de2023fighting}. %\gkso{This is possible because when dealing with non-linear models, many datasets lack identification~\cite{de2023fighting}}.
But in some situations the decision should be guided by some specificity of the system, that needs to be based on prior-knowledge of an expert~\cite{haider2022shape,martinek2024shape,poursanidis2024incorporating}.%\gkso{, or attend some constraints that are difficult or even impossible to incorporate into the objective-function}.

% Need for pattern matching
This prior-knowledge may be properties~\cite{kronberger2022shape}, unity obeying equations~\cite{reuter2024unit}, desired building blocks that must be enforced~\cite{de2022transformation} or that should not appear in the expression (see for example, the constraints that can be created in~\cite{cranmerpysr}). For example, by inspecting the data, we may notice an oscillatory behavior similar to a damped oscillator, and then it is desired that the model follow the pattern $f(x; \theta) = g(x; \theta) e^{h(x; \theta)} \cos{(k(x; \theta))}$, where $g, h, k$ represents any function. Or, by further exploring the Pareto front, we observe a bias towards a recurrent pattern such as $\tanh{(g(x))}$ and we want to explore alternatives that do not contain it, even if having to use dominated expressions not appearing in the front. %\gkso{Because of the natural bias of SR algorithms and the strictness of the dominance relationship, the final population or the Pareto front may not have enough alternative models to explore.} 
 But, integrating  prior-knowledge inside the search may not be trivial or can make the search space even harder to navigate. If we trust that the desired properties corresponds to an accurate model, we can explore the history of the visited expressions to find an accurate model that behaves as expected. This, again, creates a burden to users as it is not trivial to explore hundreds of different expressions.
 Another insightful analysis is the frequent building blocks~\cite{rosca1995towards,o1995troubling,daida:1999:aigp3,mcphee2008semantic} observed during the search that may reveal an important relationship or feature transformation that can be further inspected.

% Exploration tool
In this paper, we propose \emph{rEGGression}, an interactive tool that can help SR users to explore alternative models generated from different sources. These sources can be: the final population of a single run, the Pareto front, the entire history of visited expressions during the search, or a combination of those sources from multiple runs of the same or different algorithms. This can provide a rich library of alternative expressions generated departing from different priors (i.e., hyper-parameters, algorithms, random seeds), that can bring invaluable information about the data.
This tool supports simple queries such as querying for the top-N models filtered by size, complexity, and number of numerical parameters; insert and evaluate new expressions; list and evaluate sub-expressions of the already visited expressions, and also, more advanced queries such as calculate the frequency of common patterns (i.e., building blocks) observed in the set of models; and filter the expressions by patterns with a natural syntax.

The unique features of this tool is the \textbf{pattern matching capabilities} and the \textbf{analysis of building blocks}. The pattern matching allows to filter the current set of expressions to those that contains (or not contain) a certain functional pattern, this allows to explore models based on domain knowledge or additional assumptions that could not be incorporated during the search. The analysis of the building blocks reports the most frequent building blocks explored during the search or the building blocks with the largest average fitness. This information brings additional insights of possible patterns that contributes to obtaining a high accurate model, providing new insights about the observations. The interactive nature of this tool lies in the ability to refine the search and shifting the focus in different regions of the space of available expressions. These unique features are possible with the use of e-graph data structure~\cite{willsey2021egg} created for the equality saturation algorithm, a technique used to alleviate the phase ordering problem in the optimization of computer programs during the compilation process. 
This technique was previously used in the context of symbolic regression in~\cite{de2023reducing,kronberger2024jsc} to investigate the problem of unwanted overparameterization that increases the chance of a suboptimal fitting of the numerical parameters. The generated e-graph has another interesting feature that can be exploited by symbolic regression algorithms: it contains a database of patterns and equivalent expressions that can be easily matched against a new candidate expression.

% organization
This paper is organized as follows. In Section~\ref{sec:relatedwork} we will summarize the related works in exploration tools for symbolic regression. Section~\ref{sec:eqsat} will explain the basic concepts of equality saturation and the e-graph data structure. In Section~\ref{sec:eggGP} we will present \emph{rEGGression}, showing the available commands, and giving additional details of its inner working, specially how the pattern matching and building blocks exploration algorithms work. Finally, Section~\ref{sec:conclusions} gives some final remarks and expectations for the future.

\section{Related work}~\label{sec:relatedwork}
% Model selection, MOO, and MDL
Whenever we are confronted with multiple viable regression models with similar accuracy, we must eventually choose and commit with one of these alternatives to move forward with the analysis.
This choice is often reduced to the accuracy-simplicity trade-off, in which we seek an accurate model that can still be interpreted up to a certain level.
Model selection~\cite{stoica2004model} is part of the machine learning pipeline that deals with such a choice. Some common model selection techniques comprehends the evaluation of the model on a validation data to estimate the accuracy on unseen examples, criteria that takes simplicity into consideration such as BIC and AIC, the selection of the knee point of the Pareto front, or the recently proposed minimization of description length~\cite{bartlett2023exhaustive}. 
While these criteria help to automate the process of model selection, they cannot accommodate additional constraints based on the domain knowledge. This prompts for an interactive approach that enables an efficient exploration of the many alternative models that might be available.

% HeuristicLabs
HeuristicLab~\cite{wagner2005heuristiclab}\footnote{https://dev.heuristiclab.com/} provides a graphical interface with support to multiple search and optimization algorithms for different tasks, from nonlinear optimization to symbolic regression. Specifically for symbolic regression, it provides different variations of genetic programming and a post-processing tool of the final Pareto front (or population, if not using multi-objective) of the search displaying additional information about each model, performance plots, and a tree-based representation of the expression.
For any of these expressions, the user can prune the tree interactively to evaluate a simpler version of that model.
This tool is easy to use and adapt to many situations~\cite{elyasaf2014software}, but it limits the exploration to the final population generated by its own internal algorithm.

% DataModeler

DataModeler\footnote{https://evolved-analytics.com/}~\cite{kotanchek2013symbolic} revolves around getting insights from an ensemble of symbolic regression models generated from a single or multiple runs of their own implementation of genetic programming. They provide many different insights about the models and the data through plots and statistics about the common subset of variables shared among accurate models, commonly observed transformation of variables (i.e, sub-expressions without constant values), filtering models by the presence of common variables, just to name a few of the available supporting functionalities for data exploration.
In~\cite{kotanchek2013symbolic} the authors state:

\begin{quote}
    \textit{Since there are many factors which characterize a good model other than accuracy of fitting the observed data, the analysis tools exploited a variety of model perspectives to identify the key inputs and variable combinations which are suitable for a final model selection.}
\end{quote}
which corresponds to our view of a supporting tool for the exploration of symbolic regression models. 

% TuringBot
TuringBot\footnote{https://turingbotsoftware.com/} is another graphical interface similar to the discontinued Eureqa. It provides the Pareto front to the user with a simple report of the main characteristics of the selected model and a plot of the predictions considering a single variable.

% PySR
%PySR \cite{cranmerpysr} and PyOperon\cite{burlacu2020operon} both produce the Pareto front of the non-dominated expressions from the final GP generation.

% e-graph
% This redundancy can increase the chance of failing to correctly optimize such parameters, leading to sub-optimal solutions.
Equality saturation has been used in the context of symbolic regression as a support tool to study the behavior of the search. de Franca and Kronberger ~\cite{de2023reducing, kronberger2024jsc} show that many state-of-the-art SR algorithms have a bias towards creating expressions with redundant numerical parameters that can lead to sub-optimal results.  Simplification based on equality saturation was used to detect the equivalent expressions visited during the GP search~\cite{kronberger2024inefficiency}.
As we will explain in the next section, the e-graph data structure and supporting structures used during equality saturation can be turned into an easy to use API providing ways to search through the model space with advanced queries and pattern matching of building blocks. This can simplify the exploration of alternative expressions while following an SQL-like domain specific language.

\section{Equality saturation and e-graphs}~\label{sec:eqsat}
% phase ordering problem
During the optimization step of the compilation of computer programs,  compilers may apply a sequence of transformation rules that are known to improve the performance of the code. Some classical examples of such rules are the replacement of integer operators with bitwise operators (i.e., $2x \equiv x \ll 1$, or loop unrolling in which a \emph{for} loop with a fixed number of iterations is unrolled to avoid the \emph{jump} instruction.
The order in which such equivalence rules are applied may affect how much the program is optimized. This is known as the phase-ordering problem and demands the solution of a combinatorial optimization problem.

\text{\citet{tate2009equality}} proposed the technique called equality saturation as a solution to alleviate (in some cases, solve) the phase-ordering problem. Instead of destructively applying each equivalence rule in an arbitrary order, in equality saturation every rule is applied in parallel and the resulting equivalent programs are stored in an extension of a graph structure called \emph{e-graph}.

\begin{figure*}[t!]
    \centering
    \subfloat[]{\includegraphics[trim={5cm 16cm 2cm 1cm},clip,width=0.28\linewidth]{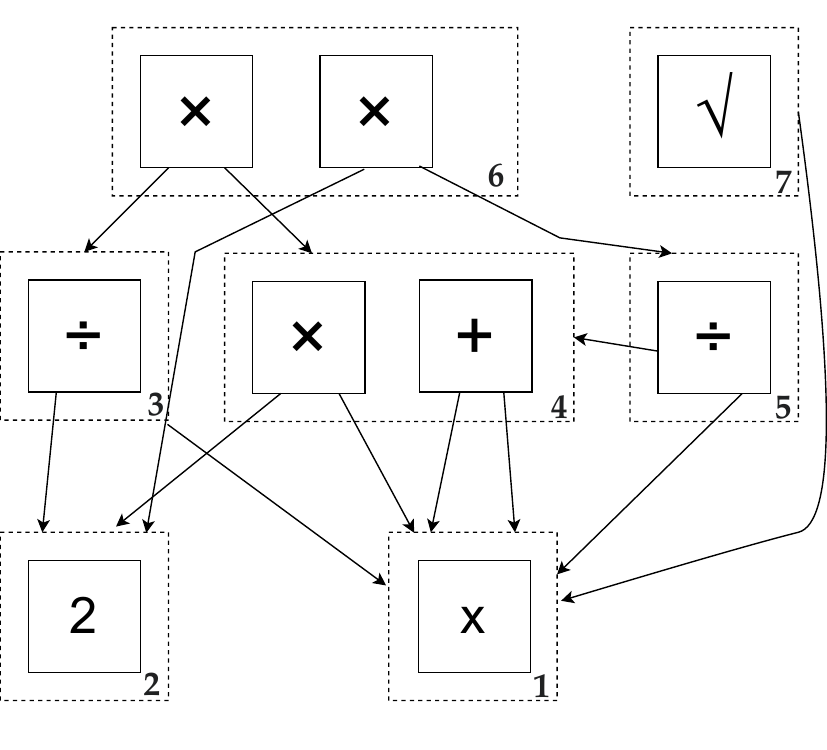}\label{fig:egraph2a}} 
    \subfloat[]{\includegraphics[trim={1cm 16cm 2cm 1cm},clip,width=0.38\linewidth]{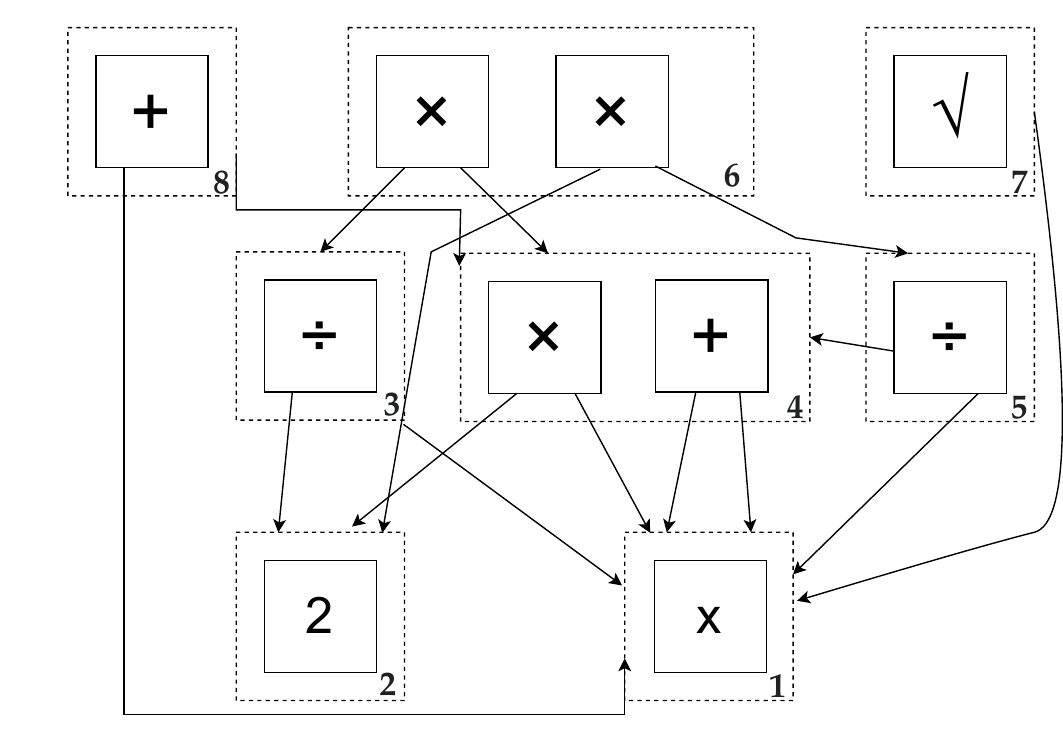}\label{fig:egraph2b}}
    \caption{(a) Illustrative example of an e-graph and (b) the same e-graph after inserting the expression $x + 2x$.}
    \label{fig:egraph2}
\end{figure*}

% e-graph with figure
An illustration of an e-graph is given in Fig.~\ref{fig:egraph2a} where each solid box represents an e-node that contains a symbol of the expression and the dashed boxes, called e-classes, groups together a set of e-nodes. Each e-class is assigned an e-class id (number on the bottom right of an e-class box). The main property of an e-class is that, all paths from any e-node contained in this e-class will generate equivalent expressions. We can see that in the middle box (e-class number $4$), if we follow through the e-node $\times$ it will generate the expression $2x$ and if we follow through the e-node $+$ it will generate $x+x$.
Conceptually, the algorithm follows four steps: 

\begin{enumerate}
    \item \textbf{Pattern matching:} Given a set of equivalence rules, search for all the e-classes matching the rule, returning a substitution map for each match.
    \item \textbf{Substitution:} Apply the rules to each match, generating a new e-class containing a single e-node.
    \item \textbf{Repair:} Merge the recently created e-class with the e-class it was generated from.
    \item \textbf{Saturate:} Repeat until saturation or a time limit.
\end{enumerate}

\begin{figure*}[t!]
    \centering
    \subfloat[]{\includegraphics[width=0.15\linewidth]{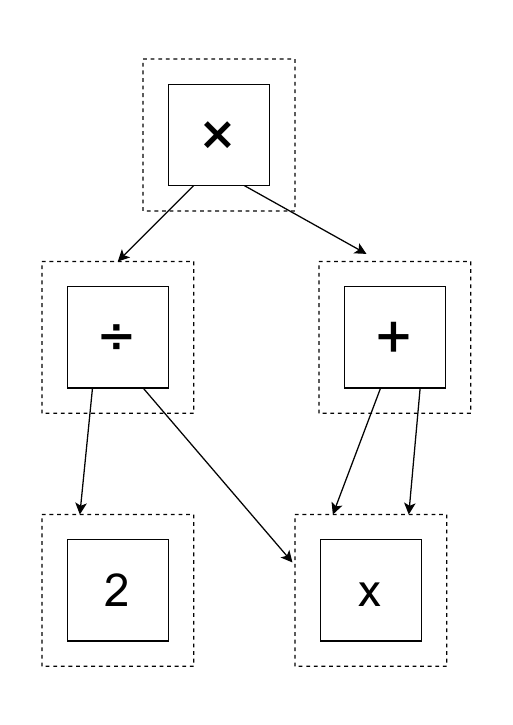}\label{fig:egg1}} \quad
    \subfloat[]{\includegraphics[width=0.19\linewidth]{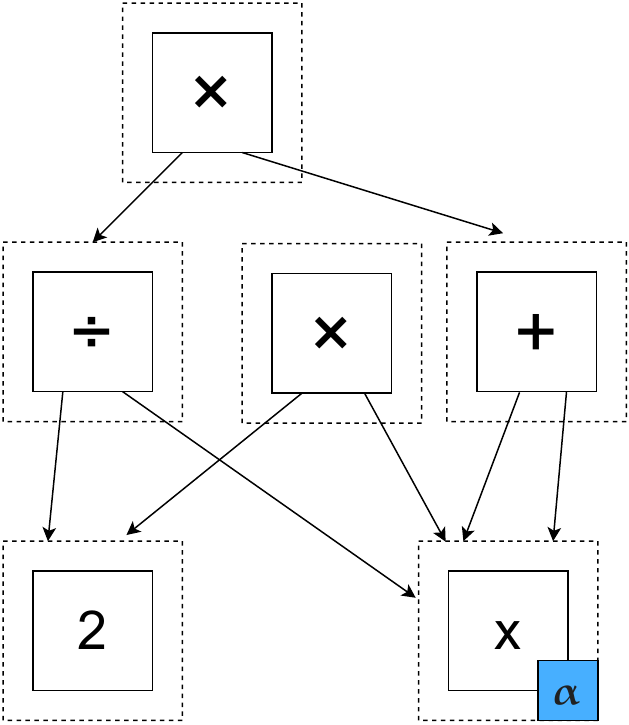}\label{fig:egg2}}  \quad
    \subfloat[]{\includegraphics[width=0.25\linewidth]{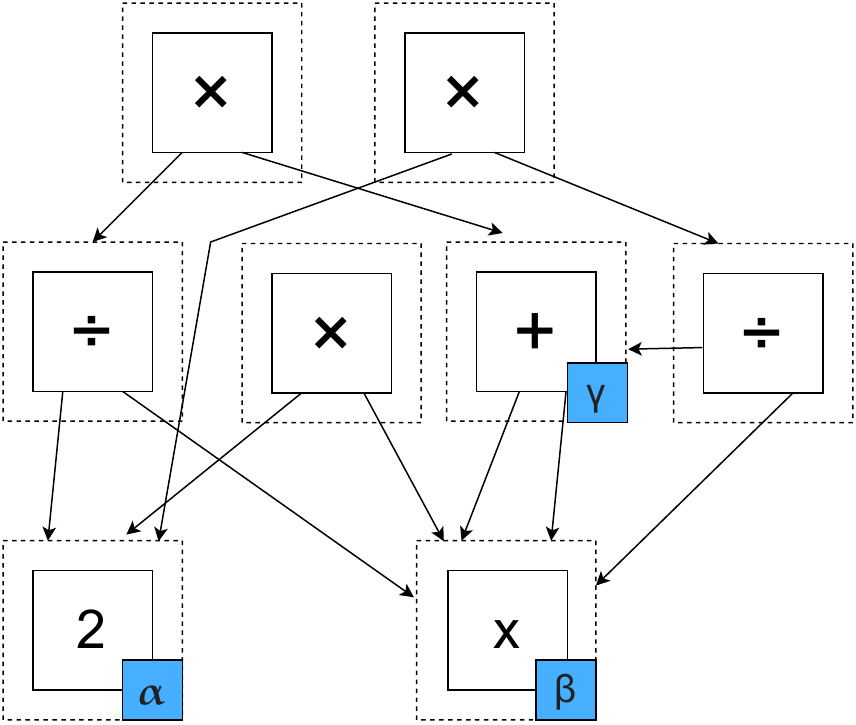}\label{fig:egg3}} \quad
    \subfloat[]{\includegraphics[width=0.25\linewidth]{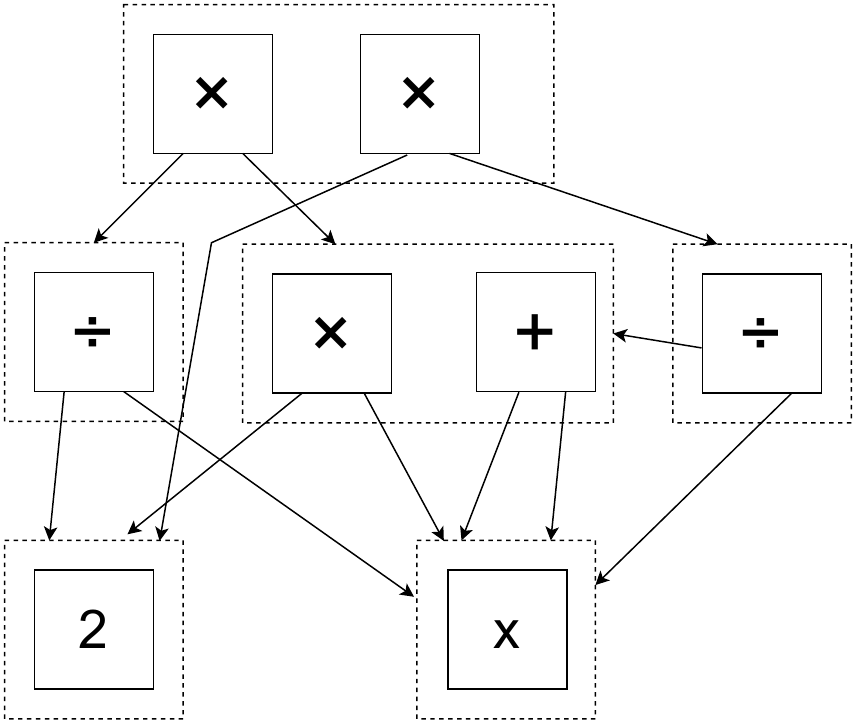}\label{fig:egg4}}
    
    \caption{Example of one iteration of equality saturation: (a) the initial e-graph after inserting the expression $(2/x)(x+x)$, (b) after matching the pattern $\alpha + \alpha$ with $\alpha=x$ and inserting the equivalent expression $2\alpha$, (c) after matching the pattern $(\alpha / \beta)\gamma$ and inserting the equivalence $\alpha (\gamma / \beta)$, (d) the final e-graph after merging the equivalent classes.}
    \label{fig:egg}
\end{figure*}

Steps $1$ to $3$ for a single equivalence rule are illustrated in Fig.~\ref{fig:egg}, where in Fig.~\ref{fig:egg1} we have the initial expression, in which we match the rules $\alpha+\alpha \equiv 2\alpha$ and $\frac{\alpha}{\beta} \gamma \equiv \alpha \frac{\gamma}{\beta}$. These rules generates two new e-classes (Figs.~\ref{fig:egg2}~and~\ref{fig:egg3}) that are merged afterwards (Fig.~\ref{fig:egg4}). Steps $1$ and $2$ are repeated for every equivalence rule and matched pattern and step $3$ is performed only after all new e-classes are recreated from the previous steps. Reaching saturation is often impossible as some rules can lead to an infinite sized e-graph.
Although the idea is simple, it requires techniques from information retrieval for an efficient implementation of the matching procedure and maintenance of the structure~\cite{willsey2021egg}. One important characteristic of this structure and the algorithm is that it will always insert new information while keeping all previous states in the same structure.

The basic data structure is composed of a map from an e-class index to an e-class index, called \emph{canonical}, with the purpose of maintaining consistency whenever we create and merge e-classes; a map from e-nodes to e-class id; a map from e-class id to the e-class data structure; and a database to optimize the pattern matching procedure. 

%In summary:

%\begin{lstlisting}[language=Haskell]
%  data EGraph = EGraph {
%    canonical : Int -> Int,
%    enodes    : ENode -> Int,
%    eclasses  : Int -> EClass,
%    db        : EGraphDB
%  }
%\end{lstlisting}

An e-node can be represented as a tree-like structure where the children point to an e-class id instead of the next token in the expression. This can be succinctly described as the tuple $(Op, [Int])$ where the first element is the $n$-ary operator, and the second element is the list of e-class ids for the children of this e-node. Since this is a \emph{hashable} structure, it can be used as a key to a map.

The e-class structure contains a set of e-nodes, a set of parent e-class ids with the corresponding e-nodes that points to the current e-class, and an extra field that contains information about the height of the e-class, the cost of the least complex expression that can be generated by that e-class, whether it evaluates to a constant value, and any additional information we may want to store (e.g., positiveness, monotonicity, convexity, and measurement units information).

\begin{figure}[t!]
    \centering
    \includegraphics[width=0.6\linewidth]{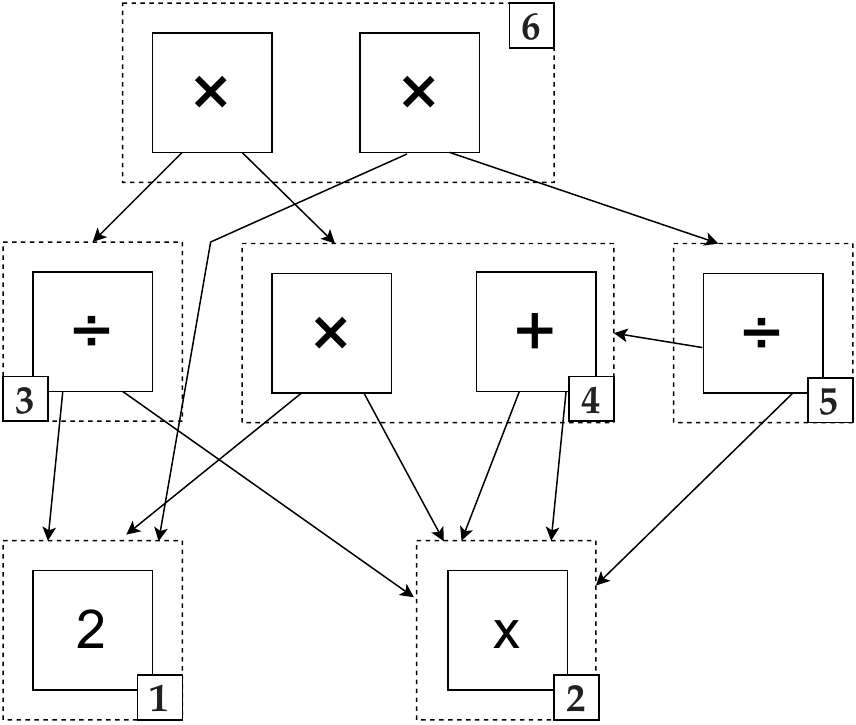}
    \caption{Reference e-graph for the pattern matching procedure example.}
    \label{fig:matching}
\end{figure}

Finally, the database stores the information of recently created e-classes that must be analyzed and merged, and a structure that stores the information of e-class ids for the smallest possible building blocks (e.g., a function or operator applied to anything). This database is a map of a token of our grammar (terminal or non-terminal) to a \emph{trie} structure~\cite{crochemore2009trie}. This \emph{trie} contains the information of all e-classes and their children containing that particular token. This allows us to retrieve all e-classes that have at least one e-node with a certain token and their corresponding children.
Considering the e-class ids of Fig.~\ref{fig:matching} (small numbered boxes), querying the token $\times$ from the database will return a \emph{trie} with the keys $4, 6$, corresponding to the e-classes containing that token. The key $4$ will point to a \emph{trie} with a single key $1$ which points to the key $2$. The key $6$ will point to the keys $1$ and $3$ corresponding to each e-node's first child, and they will respectively point to the keys $4$ and $5$. 

%\begin{lstlisting}[language=Haskell]
%    data DB   = Token -> Trie
%    data Trie = Int -> Trie
%\end{lstlisting}

As an illustrative example, consider the query for the pattern $\alpha \times (\beta + \gamma)$. In the first step, it creates an empty list of substitution maps of tokens and variables to e-class ids:
\begin{align*}
    \{ \times \to ?, \alpha \to ?, + \to ?, \beta \to ?, \gamma \to ? \} 
\end{align*}

Next, it retrieves the e-class ids for the token $\times$, which are $4, 6$, creating two maps, one for each key:
\begin{align*}
    \{ \times \to 4, \alpha \to ?, + \to ?, \beta \to ?, \gamma \to ? \} \\
    \{ \times \to 6, \alpha \to ?, + \to ?, \beta \to ?, \gamma \to ? \}
\end{align*}

Following up the \emph{trie} for the ids of the first child, it replaces the values of $\alpha$:
\begin{align*}
    \{ \times \to 4, \alpha \to 1, + \to ?, \beta \to ?, \gamma \to ? \} \\
    \{ \times \to 6, \alpha \to 3, + \to ?, \beta \to ?, \gamma \to ? \} \\
    \{ \times \to 6, \alpha \to 1, + \to ?, \beta \to ?, \gamma \to ? \}
\end{align*}

Since $\alpha$ is a pattern variable, we do not need to check whether the mapping is correct. The next entry for each \emph{trie} will fill up the values for the $+$ token:
\begin{align*}
    \{ \times \to 4, \alpha \to 1, + \to 2, \beta \to ?, \gamma \to ? \} \\
    \{ \times \to 6, \alpha \to 3, + \to 4, \beta \to ?, \gamma \to ? \} \\
    \{ \times \to 6, \alpha \to 1, + \to 5, \beta \to ?, \gamma \to ? \}
\end{align*}

Since $+$ is not a variable, it must be checked against the entry in the database, which points to the set $\{4\}$. By intersecting with the three maps we have, it will prune the infeasible maps and lead to:
\begin{align*}
    \{ \times \to 6, \alpha \to 3, + \to 4, \beta \to ?, \gamma \to ? \} 
\end{align*}

Since both children are pattern variables, the final map will be:
\begin{align*}
    \{ \times \to 6, \alpha \to 3, + \to 4, \beta \to 2, \gamma \to 2 \} 
\end{align*}

%Notice that if our pattern was $\alpha * (\beta + \beta)$, in the final step we would have
%\begin{align*}
%    \{ \times \to 6, \alpha \to 3, + \to 4, \beta \to 2 \} 
%\end{align*}
%after checking whether that both children of $+$ points to $2$.

\begin{figure}[t!]
    \centering
    \subfloat[]{\includegraphics[width=0.70\linewidth]{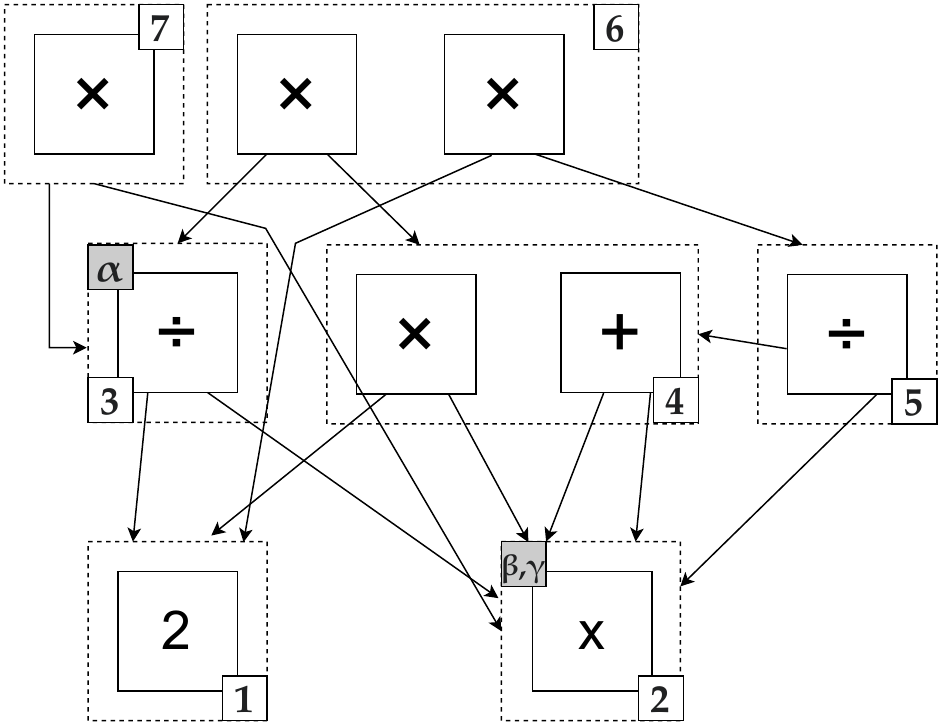}\label{fig:insertion1}} \\
    \subfloat[]{\includegraphics[width=0.70\linewidth]{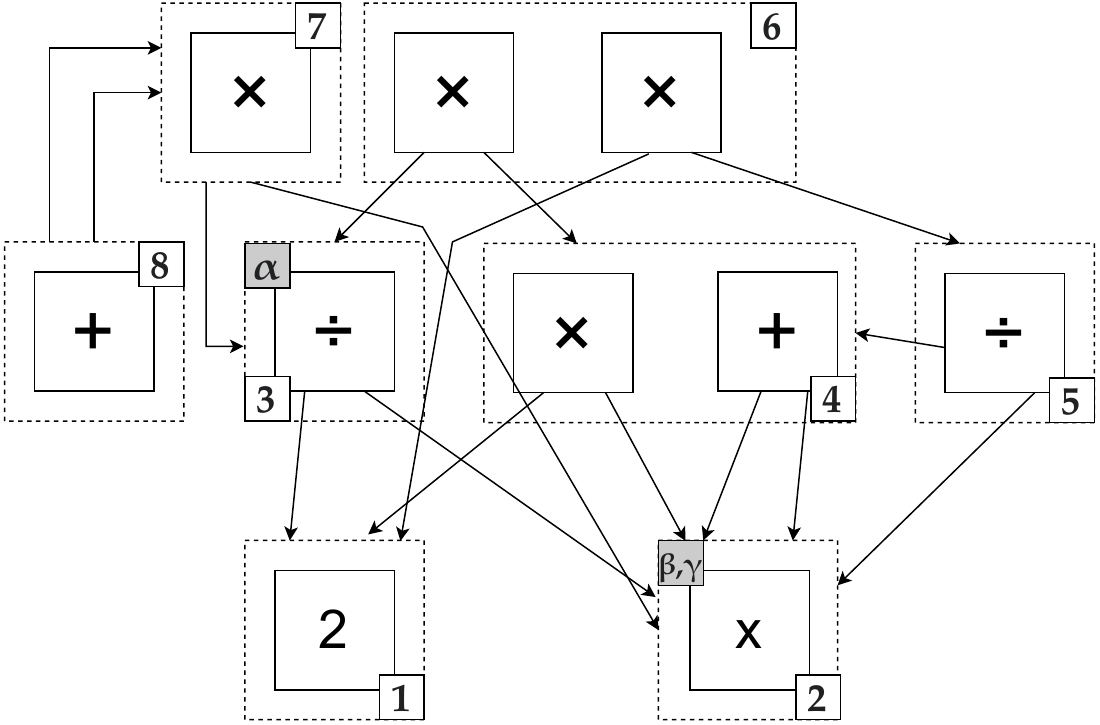}\label{fig:insertion2}} 
    \caption{Inserting the expression after matching.}
    \label{fig:insertion}
\end{figure}

After the final step, this procedure will return a set of substitution maps that will point to each match of that pattern and can be used to create equivalent expressions. Suppose we want to create the equivalent expression $(\alpha \times \beta) + (\alpha \times \gamma)$, we start by inserting $(\alpha \times \beta)$, or the e-node $(\times, [3, 2])$ (Fig.\ref{fig:insertion1}), which would generate a new e-class with id $7$, we then insert the e-node corresponding to the right side of the expression, $(\times, [3, 2])$, which would simply return the e-class id $7$ (i.e., it will not create the same expression again), and we would finally create the e-node $(+, [7,7])$, which would prompt the creation of a new e-class id $8$ (Fig.\ref{fig:insertion2}). This e-class would be marked to be merged with the e-class id $6$ as they are equivalent. 

Regarding the complexity of these operations, the building of the database is linear, while the querying is bounded by $O(\sqrt{q n^m})$, where $q$ is the number of returned queries, $n$ is the number of e-nodes in the e-graph, and $m$ is the number of conjunctive queries, proportional to the number of non-terminal nodes. 
More details of an efficient implementation of e-graph and equality saturation can be found in~\cite{willsey2021egg} and for the pattern matching algorithm in~\cite{zhang2021relational}.

%\begin{figure}[t!]
%    \centering
%    \includegraphics[trim={5cm 16cm 2cm 1cm},clip,width=0.5\linewidth]{figs/egraph-full.pdf}
%    \caption{Example of an e-graph.}
%    \label{fig:egraph}
%\end{figure}

% structure details
%There are currently a few implementations of equality saturation, notably \emph{egg}\footnote{https://docs.rs/egg/latest/egg/ egg stands for \textbf{e}-\textbf{g}raphs \textbf{g}ood} in Rust~\cite{willsey2021egg}, \emph{hegg}\footnote{https://github.com/alt-romes/hegg/} in Haskell, \emph{Metatheory.jl}\footnote{https://juliasymbolics.github.io/Metatheory.jl/dev/egraphs/} in Julia.
% TODO: uncoment after review
% and \emph{srtree}~\footnote{https://github.com/folivetti/srtree}, also in Haskell but specifically crafted for symbolic regression.

\section{rEGGression: an exploration tool for symbolic regression models}~\label{sec:eggGP}
In this paper, we are proposing a new software, called \emph{rEGGression},  an exploration tool for symbolic regression models that uses e-graphs to perform advanced queries. This tool is compatible with the output of many popular symbolic regression implementations (see Sec.\ref{sec:compat}).
The key highlights of rEGGression are:

\begin{enumerate}
    \item supports querying for expressions containing or not containing specific patterns.
    \item calculates the distribution of building blocks that can be later used for querying.
    \item it supports importing expressions from many popular symbolic regression implementations.
\end{enumerate}

The implementation of rEGGression follows closely the implementation of e-graph and equality saturation described in~\cite{willsey2021egg} and the pattern matching in~\cite{zhang2021relational}. The name of this tool is a direct reference to the current main e-graph library \emph{egg} written in Rust\footnote{https://egraphs-good.github.io/}.

\subsection{Available commands}
The available commands of rEGGression are:

\begin{itemize}[leftmargin=*]
    \item \textbf{top}: returns the top $N$ expressions following a criterion.
    \item \textbf{report}: reports details about the expression with a given id.
    \item \textbf{subtrees}: return the ids and information about the subtrees of a certain expression.
    \item \textbf{optimize}: (re)optimizes a certain expression by minimizing the specified loss function.
    \item \textbf{insert}: inserts, evaluates and displays the information of a new expression.
    \item \textbf{pareto}: shows the Pareto front where the first objective is either the fitness or the description length, and the second objective is the model size.
    \item \textbf{count-pattern}: counts the occurrences of a building block.
    \item \textbf{distribution}: shows the distribution of all observed building blocks up to a certain size (limited up to size $10$).
    \item \textbf{save, load}: saves or loads the current state of the e-graph.
    \item \textbf{import}: import the expressions from a CSV file.
\end{itemize}

In the following, all these commands will be demonstrated using the \emph{nasa\_battery\_1} dataset containing observations of the current state of a battery and its current capacity.
The e-graph was generated with \emph{eggp}\cite{eggp} algorithm that makes use of e-graph internally. The hyper-parameter values were: MSE loss function, $200$ generations, population size of $100$, $+,-,\times,\div,\sin,e,\log,|.|,\operatorname{pow}$ non-terminals set, maximum size of the expression of $20$.
% To start the software, we run the following command line in the console application:

%\begin{verbatim}
%reggression -d nasa_battery_1_10min.csv \
%  --load-from nasa_battery.egraph  \
%  --distribution MSE
%\end{verbatim}

\subsection{Showing the top matching expressions}
In our implementation, we keep two sorted lists of e-class ids, one by fitness and the other by description length (dl). These lists use a fingertree~\cite{hinze2006finger} data structure to allow a $O(\log n)$ insertion and deletion, but this can be implemented with any structure with this asymptotic complexity. The command \emph{top} follows the format:

\begin{verbatim}
top N [ FILTER... ] [ CRITERION ] 
  [ [not] matching [root] PATTERN ]

FILTER: with ( size | cost | parameters )
             ( < | <= | = | > | >= ) N 
CRITERION: by fitness | by dl
\end{verbatim}
where \emph{[ARG]} represents optional arguments and \emph{...} means that that argument can be repeated. For example, running the command \emph{top 3 with size < 5}, will result in:

{\small
\begin{tabular}{cccccc}
\toprule 
    Id & Expression & Fitness & Parameters & Size & DL \\
     \midrule
     $1622$ & $(t0 / x3)$ & $-36990.9065$ & $[569.9222]$ & $3$ & -- \\
     $5060$ & $(t0 / \sin(x3))$ & $-39292.8520$ & $[558.7884]$ & $4$ & -- \\
     $10220$ & $(\log(x3) * t0)$ & $-40354.3007$ & $[-1563.021]$ & $4$ & -- \\
     \bottomrule
\end{tabular}
}

The first column is the id of the expression that can be referred to in other commands, the second column is the expression itself with the variables represented by $x_i$ and adjustable parameters as $t_i$. The third column is the fitness (assuming maximization) assigned to that expression by the algorithm, the fourth column is the values of the parameters, followed by the number of nodes and the DL, when available. The DL can be calculated when starting the program with the flag \emph{--calculate-dl}. Alternatively, this can be calculated on-demand with the \emph{report} command (demonstrated on a later section).
The filter argument can be composed by adding new \emph{with} directives, such as:

\begin{itemize}[leftmargin=*]
    \item \emph{top 3 with size > 3 with size < 6 with parameters > 1}: it will retrieve the top $3$ expressions with size between $3$ and $6$ non inclusive and with more than one parameter.
    \item \emph{top 3 with cost < 10}: will retrieve the top $3$ constrained by the default cost function. The cost function adds up an assigned values to each node ($1$ for constants, variables, and parameters, $2$ for binary operators and $3$ for unary operators). In a future version, the user will be able to customize this cost function.
    \item \emph{top 3 with size < 6 by dl}: retrieves the top $3$ with size less than $6$ sorted by description length.
    \item \emph{top 3 matching v0 + v0}: retrieves the top $3$ expressions that matches the specific pattern.
\end{itemize}

The \emph{top} command will parse the filter options and create a predicate function. It will then traverse the sorted list while keeping the elements that returns true to the predicate until the requested number of expressions is reached. This has a complexity of $O(n)$ in the worst case considering a total of $n$ expressions in the database.

\subsubsection{Pattern matching}
The pattern matching has two optional arguments: \emph{root} and \emph{not}, where the first tells the program to only match expressions following the pattern at its root while the second retrieves the complement of the pattern matching result. For example, the expression $x_0 * (x_1 + \theta_0)$ would match the pattern $(v_0 + v_1)$, but it would not match it if the argument \emph{root} is enabled. The pattern matching algorithm works as described in Sec.~\ref{sec:eqsat}, returning the list of matches at the root. If the \emph{root} flag is disabled (default), it will also recursively retrieve all the parents of each matched expression. After returning the list of all e-classes, we traverse the sorted list of expressions while applying the filter predicate and check whether the e-class id belongs (or not, depending on the argument) to the list of matched e-classes.

The syntax for the pattern follows the same syntax of a mathematical expression, where by our convention $x_i$ is the $i$-th variable, starting from $0$, $t_i$ is the $i$-th adjustable parameter, and $v_i$ is a pattern variable, that matches everything. If we use a repeated index for the pattern variable, it will ensure that they both match the same expression. So, the pattern $v_0 + v_0$ will match $(x_0 + 2) + (x_0 + 2)$ or $(x_0 + 2) + (2 + x_0)$, but not $(x_0 + 2) + (x_0 \times 2)$.

\subsection{Detailed report}
With the command \emph{report id}, the program display a more detailed report of the expression with this e-class \emph{id}.
Besides the same information of the \emph{top} command, it will also show the MSE, $R^2$, negative log-likelihood, and DL for the training and test datasets, if a test set is provided. Additionally, the DL of that expression will be stored in the database if previously unavailable.

\subsection{Disassembling the expression tree}
The command \emph{subtrees id} will display all the subtrees of the expression with that id. It will display the same information as the \emph{top} command, if available. This command is useful when you want to explore the building blocks of a certain expression. In our example, the expression $(\theta_0 x_3) + (\theta_1 x_1)$, will be deconstructed as: $\theta_0 x_3$, $\theta_0 x_1$, $\theta_0$, $x_1$, $x_3$. With the provided information, we can check whether one of these subtrees have a better fitness than (or close to) the whole expression, in which case we may prefer the simpler solution.

%Additionally, by observing the subtrees of two expressions, we may want to test a combination of those to test a new expression our own. This can be done with the \emph{insert} command.

\subsection{Inserting new expressions}
Using the \emph{insert} command we can test expressions that were not previously explored and insert them into the database. By observing the subtrees of the top $2$ expressions, we verify that the subtrees $\theta_0 \sqrt{x_0}$ and $\theta_0 x_4$ both have a reasonable fitness. If we insert the expression with the command \emph{insert t0 * sqrt(x0) + t0 * x4}, we obtain a new expression with a better fitness than both subtrees.

\subsection{Optimizing expression parameters}
Depending on the expression, the optimization of the coefficients may induce a multimodal optimization problem, in such a case it is advisable to try optimizing multiple times with different starting points to try to find a better optimum. For example, inserting a new expression with the command \emph{insert sqrt(x0 |**| t0)}, where \emph{|**|} represents the protected power operator $|x|^y$, if we run the command \emph{optimize} multiple times we get a fitness value ranging from $-2\,853\,163$ to $-1\,616\,727$.

\subsection{Pareto front}
The command \emph{pareto (by fitness| by dl)} will display the Pareto front where the first objective is either the fitness value or the description length, and the second objective is the size of the expression. The output format is the same of the command \emph{top}.
To make the retrieval of the front efficient, we exploit the fact that the second objective, size, is a discrete objective with limited values. In the implementation, we created an array with the size equal to the maximum observed size and each element points to two sorted lists, one with the expressions of that particular size sorted by fitness and the other sorted by DL. Since we only store the e-class ids as integers, this can be made memory efficient allowing a fast retrieval of the Pareto front and subsequent fronts.

\subsection{Counting building blocks}
The command \emph{count-pattern} returns the number of expressions containing the desired pattern (expressions and sub-expressions). The algorithm is the same as the pattern matching in the \emph{top} command, but returns the number of e-class ids found during the matching.
In our example database, the pattern $v_0 + v_1$ is observed $103\,927$ times inside the expressions, while the pattern $v_0 + v_0$ is contained in $99\,009$ expressions. A larger building block represented by the pattern  $v_0 + sin(t_0 + x_0)$ appears in only $30$ expressions.

\subsection{Distribution of building blocks}
Searching for an specific building block can be insightful if we already have some prior-knowledge about the expected shape of the function. Alternatively, it is also of interest to have a list of the common building blocks observed during the search. 
For this purpose, rEGGression provides the command \emph{distribution [ with size ( < | <= | = | > | >= ) m ] [limited at n] (by count | by fitness) [with at least x] [from top y]}. 

This command will return a report of the building blocks extracted from the top $y$ evaluated expressions, with size constrained by $m$ and a comparison operator, limited at the $n$ first sorted by either the frequency or fitness, filtering the list to those patterns with a frequency of at least $x$.

For example, when issuing the command \emph{distribution with size <= 7 limited at 25 by fitness with at least 1000 from top 10000}, we obtain Table~\ref{tab:buildingblocks} as a result, listing the top $25$ patterns with size less than or equal to $7$, sorted by the average fitness with a frequency of at least $1\,000$ extracted from the top-$10000$ expressions.
We should notice that the extraction of building blocks is independent of the equality saturation procedure, thus it will count patterns such as $v_0 + (v_1 + v_2)$ and $(v_0 + v_1) + v_2$ as two distinct patterns.

\begin{table}[t!]
\centering
\caption{Count and average fitness of the top-$25$ building blocks sorted by average fitness.}\label{tab:buildingblocks}
\begin{tabular}{ccc}
  \toprule
  Pattern & Count & Avg. Fitness \\
  \midrule
    (v0 + (v1 + (v2 + v3)))                            &  2004 &    -4.3597e3 \\
 (((v0 / v1) + v2) + v3)                            &  1156 &    -4.3788e3 \\
 ((v0 + v1) + (v2 + v3))                            &  1527 &    -4.4102e3 \\
 ((v0 + (v1 + v2)) + v3)                            &  1380 &    -4.4143e3 \\
 ((v0 * v1) + (v2 - v3))                            &  1359 &    -4.4478e3 \\
 (v0 + ((v1 + v2) + v3))                            &  2034 &    -4.4857e3 \\
 (v0 + (v1 - (v2 + v3)))                            &  1739 &    -4.4908e3 \\
 ((v0 + v1) + v2)                                   &  3966 &    -4.4979e3 \\
 (v0 + (v1 + v2))                                   &  4735 &    -4.5203e3 \\
 (v0 + (v1 * v2))                                   &  1358 &    -4.5484e3 \\
 (v0 + (v1 - v2))                                   &  2177 &    -4.5623e3 \\
 ((v0 * v1) + v2)                                   &  2507 &    -4.5706e3 \\
 (v0 + v1)                                          &  9976 &    -4.5981e3 \\
 ((v0 * v1) + (v2 + v3))                            &  1069 &    -4.7027e3 \\
\bottomrule
\end{tabular}
\end{table}

Using this table, we can then match the expressions with the \emph{top} command and test new combinations of building blocks with \emph{insert}.

Notice that it is important to establish a limit on the pattern size and number of expressions because of the fast growing nature of the possible patterns for each expression.
All the building blocks of an expression can be extracted with a recursive procedure. A leaf node will contain two building blocks, a pattern variable $v$, that matches everything, and the node itself (e.g., $x$). An unary node will generate the building blocks $v$ and all the building blocks extracted from its child after applying this unary operator. A binary node will generate the build block $v$ together with all the building blocks created after applying the operator to the combinations of building blocks of the left and right children. Considering any expression represented as a complete binary tree, the number of building blocks can be calculated as:
\begin{align*}
    f(1) &= 2 \\
    f(n) &= 1 + f(n-1)^2
\end{align*}
where $n$ is the number of levels of that tree. Table~\ref{tab:levels} shows the number of building blocks for trees with up to $5$ levels ($32$ nodes).

\begin{table}[th!]
\centering
\caption{Number of possible patterns for tree of different number of levels.}\label{tab:levels}
\begin{tabular}{cc}
\toprule
Level  & Number of patterns \\
\midrule
    $1$ & $2$ \\
    $2$ & $5$ \\
    $3$ & $26$ \\ 
    $4$ & $677$ \\
    $5$ & $458\,330$ \\
    \midrule
    Total & $459\,040$ \\
    \bottomrule
\end{tabular}
\end{table}

As we can see, depending on the number of expressions and the maximum size, these numbers will quickly grow to a number of building blocks that will not fit in memory.
To generate the pattern, we first select the top $y$ expressions, and for each of them, we extract all the patterns to a map structure mapping the pattern to a tuple of the frequency of that pattern and the sum of the fitness. We combine all the generated maps by adding the values and, afterwards, calculating the average fitness by diving the sum of fitness with the frequency of that pattern. The algorithm used to extract the counts for a single expression is illustrated in Alg.~\ref{alg:pattern}. In our implementation, each node is represented as an e-node, so the children points to the next e-class. This can be used to optimize the step where the current node represent a binary operator since we can check whether both left and right sides are the same. In this situation, we do not need to call the main function twice.

\begin{algorithm}[t!]
    \caption{Algorithm to extract all the building blocks of an expression. In this code $v$ represents a pattern variable.}\label{alg:pattern}
    \begin{algorithmic}[1]
        \Procedure{getPatterns}{$n$} \Comment{Return the building blocks of root node $n$.}
        \If{$\operatorname{isTerminal}(n)$}
          \State $new\_pats \leftarrow [(n,1)]$
        \EndIf
        \If{$\operatorname{isUni}(n)$}
           \State $(f, t) \leftarrow n$
           \State $patterns \leftarrow \operatorname{getPatterns}(t)$
           \State $new\_pats \leftarrow [f(pat)\; |\; pat \leftarrow patterns]$
        \EndIf
        \If{$\operatorname{isBin}(n)$}
           \State $(op, l, r) \leftarrow n$
           \State $l\_pats \leftarrow \operatorname{getPatterns}(l)$
           \State $r\_pats \leftarrow \operatorname{getPatterns}(r)$
           \State $new\_pats \leftarrow [op(lp, rp)\; |\; lp \leftarrow l\_pats,\, rp \leftarrow r\_pats]$
        \EndIf
        \State \textbf{return} $[(v, 1)] + new\_pats$
        \EndProcedure
    \end{algorithmic}
\end{algorithm}

\subsection{Saving, loading, and importing}\label{sec:compat}
During the exploration of the e-graph, we may wish to save and later load any insertion we made during the exploration. The software supports the commands \emph{save} and \emph{load} that will save the e-graph to a binary format and load a saved e-graph. Notice that since this command stores the whole data structure, it may require a large number of bytes. Our example e-graph, created by evaluating $100\,000$ expressions, generated a $200MB$ file. This size can vary depending on how much redundancy there is in the sub-expressions, which will lead to a more compact representation.

It is also possible to import expressions generated by different symbolic regression algorithms with the command \emph{import filename True~|~False} that will import the expressions stored in the file \emph{filename}. The boolean value indicates whether the parser needs to extract the parameter values from the expression (\emph{True}) or if they are already provided in the file (\emph{False}). The extension of the filename will tell the parser  which algorithm generated the expressions. For example, for the filename \emph{egraph.operon}, it will be assumed that the expressions follow the same format as Operon~\cite{burlacu2020operon}. The supported algorithms are: Operon~\cite{burlacu2020operon}, Bingo~\cite{randall2022bingo}, TIR~\cite{de2022transformation}, ITEA~\cite{de2021interaction}, GOMEA~\cite{virgolin2017scalable}, SBP~\cite{virgolin2021improving}, EPLEX~\cite{la2016epsilon}, FEAT~\cite{la2018learning}, PySR~\cite{cranmerpysr}, HeuristicLab (HL)~\cite{wagner2005heuristiclab}.

The imported expressions are inserted into the current e-graph, effectively increasing the number of expressions. This can be used to combine the results from different runs of the same or distinct algorithms.
We should notice that many other implementations follows at least one of the formats used by one of these algorithms in the list. More algorithms will be supported on demand. The imported file must follow a comma separated value format with the expression in the first column, a semicolon separated list of coefficient values in the second column, and the fitness value in the third column. For example: \verb|x0^p0 + p1*x1,0.2;3.1,0.89|.

\iffalse 
\subsection{Command line}
The command line used to start rEGGression offers different choices to customize the initial experience. The available arguments are:

\begin{itemize}
    \item \emph{--calculate-dl}: this argument will calculate the DL for each one of the e-class ids that already contains a fitness value.
    \item \emph{--convert filename}: converts the equations in \emph{filename} to a standard format (TIR).
    \item \emph{--dataset filename}: the name of the training dataset used to generate the equations.
    \item \emph{-t filename}: a dataset to be used as the test set in \emph{report} command.
    \item \emph{--distribution MSE|Gaussian|Poisson|Bernoulli|ROXY}: the distribution or error metric used to calculate the fitness function. Notice that for Gaussian, the parameter vector must contain the value of $\sigma$ as the last value, for ROXY~\cite{bartlett2023marginalised}, the dataset must contain the error columns.
    \item \emph{--parse-csv}: load the expressions from a CSV file instead of an e-graph file.
    \item \emph{--parse-parameters}: whether to parse the parameters from the expression or use the second column as the parameter values.
\end{itemize}

We should notice that the user must ensure the correct choice of arguments reflecting the fitness values during the symbolic regression search. Mixing expressions and fitness values from different error functions will make the exploration inconsistent to reality.
\fi 

\subsection{Availability}

The source files and compiled binaries for Linux, MacOS, and Windows are available at the \emph{srtree} symbolic regression library\footnote{\url{https://github.com/folivetti/srtree/}} repository. The program requires that the NLOpt library~\cite{johnson2021nlopt} is correctly installed and accessible in the current system.
%The files used for this experiment is available at \url{FILL UP}.
The \emph{srtree} library implements the e-graph data structure and equality saturation algorithms with an extensive set of support functions that can make it easy to extend \emph{rEGGression} with new functionalities.

\subsection{Advantages and limitations}
The rEGGression tool is not meant to be a replacement to the other available tools, like the ones mentioned in Sec.\ref{sec:relatedwork}. This work is meant to offer the exploration of a large selection of regression models beyond the common tradeoff between accuracy and complexity, allowing the user to refine the search towards a model that is compatible with the expectations.
The most distinct features of rEGGression, when compared to other tools are: the easy-to-use pattern matching of expressions and the reporting of common building blocks. 
With the pattern matching capability, the user can refine the current selection of top models by enforcing or prohibiting certain building blocks. Likewise, reporting the common building blocks reveals the recurring patterns that may be responsible for the high accuracy of the top models, this can bring additional insights for explaining the behavior of the observations. Finally, in future versions, we will store additional information about the expressions such as whether they are positive, monotonic, and their units when this information is available. This information could be used as additional constraints to the selection of top expressions.

As the main limitations of rEGGression, first of all it is limited to a command line interface. Although the commands are very straightforward, as we add new functionalities it can become increasingly difficult to use. Second, although it supports symbolic expressions as exported by many SR algorithms, it is not possible to integrate this tool into a programmable pipeline (e.g., a Python script that process the data, run the algorithm, and perform post-analysis with rEGGression). Finally, some of the reports could be presented in the form of plots, such as the Pareto front.

\section{Conclusions}~\label{sec:conclusions}
In this paper we present rEGGression, a tool with the main objective of exploring symbolic regression models generated by multiple sources. The main idea of this tool is to investigate a large database of SR models to explore different alternatives than those selected by the search algorithm. The tool brings a text-based interface supporting an SQL-like language in which the user can query the top expressions, search for patterns, and get a report of common building blocks.

The backend of this tool uses the e-graph data structure, commonly used in the simplification and optimization of computer programs. This data structure allows an efficient implementation of a pattern matching system of building blocks in the stored expressions.

With the help of this data structure, rEGGression supports different commands such as displaying the top $N$ expressions, sorted by either the fitness or the description length, search and retrieve the top expression that contains (or not contain) a certain pattern, dissect an expression into their subtrees, insert new expressions, reevaluate an expression, display the Pareto front, count the occurrence of a certain pattern, and report the common building blocks given a filter criteria.

From this set of commands, the main highlight is the exploration of the building blocks which can provide frequent patterns observed in expressions with good accuracy, and the average fitness associated with those patterns. This can help users interact with the database of expressions, allowing them to test new hypotheses combining accurate expressions or exploiting the building blocks. 

There are many possibilities for future versions of this software. One of them is concerned to the amount of information obtained from a single expression using the \emph{report} command. Besides the information already shown in this tool, we can also display the expression as a function of observed transformed variables, for example the expression $\theta_0 sin(x^2) + \theta_1 e^{-x}$ can be displayed as $\theta_0 z_0 + \theta_1 z_1$, which allows us to treat this as a linear model. Besides that, we could also calculate and display the confidence interval of the parameters, and make single point predictions with the confidence intervals.% correlation between parameters.%, and the estimated effect of each variable using mean partial effect or partial effect at the mean. 
We will also add a \emph{simplify} command that will run equality saturation on a certain expression to try to reduce its code length.

The e-graph can also be used to infer properties about the expression  and its parts such as units of measure~\cite{reuter2024unit} or bounds for the output domain via semantic analysis rules. By doing this, we can filter the expression according to the desired properties.
%\gkso{such as positivity, monotonicity, concavity, symmetry, etc. We just simply need to model the rules that ensure these properties (i.e., $x^2$ is guaranteed to be positive) and perform the analysis when building the e-graph. Also, we can incorporate unity information and propagate the units to each e-class of the graph}.

\begin{acks}
F.O.F. is supported by Conselho Nacional de Desenvolvimento Cient\'{i}fico e Tecnol\'{o}gico (CNPq) grant 301596/2022-0.

G.K. is supported by the Austrian Federal Ministry for Climate Action, Environment, Energy, Mobility, Innovation and Technology, the Federal Ministry for Labour and Economy, and the regional government of Upper Austria within the COMET project ProMetHeus (904919) supported by the Austrian Research Promotion Agency (FFG). 
\end{acks}

\bibliographystyle{ACM-Reference-Format}
\bibliography{references}

\iffalse
\appendix

\section{Online Resources}

All data and the source code for implementations, experiments, and post-processing analysis are available at \url{https://github.com/}.
\fi

\end{document}